\documentclass[nohyperref]{article}

\usepackage{microtype}
\usepackage{graphicx}
\usepackage{subfigure}
\usepackage{booktabs} 

\usepackage{hyperref}

\usepackage[accepted]{icml2022}

\usepackage{amsmath}
\usepackage{amssymb}
\usepackage{mathtools}
\usepackage{amsthm}

\usepackage[capitalize,noabbrev]{cleveref}

\theoremstyle{plain}

\theoremstyle{definition}

\theoremstyle{remark}

\usepackage[textsize=tiny]{todonotes}

\icmltitlerunning{An Adaptive Deep Clustering Pipeline to Inform Text Labeling}

\begin{document}

\twocolumn[
\icmltitle{An Adaptive Deep Clustering Pipeline to Inform Text Labeling at Scale}



\icmlsetsymbol{equal}{*}

\begin{icmlauthorlist}
\icmlauthor{Xinyu Chen}{equal,verint}
\icmlauthor{Ian Beaver}{equal,verint}
\end{icmlauthorlist}

\icmlaffiliation{verint}{Verint Systems Inc., Melville NY, USA}

\icmlcorrespondingauthor{Xinyu Chen}{Xinyu.Chen@verint.com}
\icmlcorrespondingauthor{Ian Beaver}{Ian.Beaver@verint.com}

\icmlkeywords{Machine Learning, ICML}

\vskip 0.3in
]



\printAffiliationsAndNotice{\icmlEqualContribution} 

\begin{abstract}
Mining the latent intentions from large volumes of natural language inputs is a key step to help data analysts design and refine Intelligent Virtual Assistants (IVAs) for customer service and sales support. We created a flexible and scalable clustering pipeline within the Verint Intent Manager (VIM) that integrates the fine-tuning of language models, a high performing k-NN library and community detection techniques to help analysts quickly surface and organize relevant user intentions from conversational texts.  The fine-tuning step is necessary because pre-trained language models cannot encode texts to efficiently surface particular clustering structures when the target texts are from an unseen domain or the clustering task is not topic detection.  We describe the pipeline and demonstrate its performance and ability to scale on three real-world text mining tasks.  As deployed in the VIM application, this clustering pipeline produces high quality results, improving the performance of data analysts and reducing the time it takes to surface intentions from customer service data, thereby reducing the time it takes to build and deploy IVAs in new domains.
\end{abstract}

\begin{figure*}
\hspace{0.5cm}
  \includegraphics[width=0.9\textwidth]{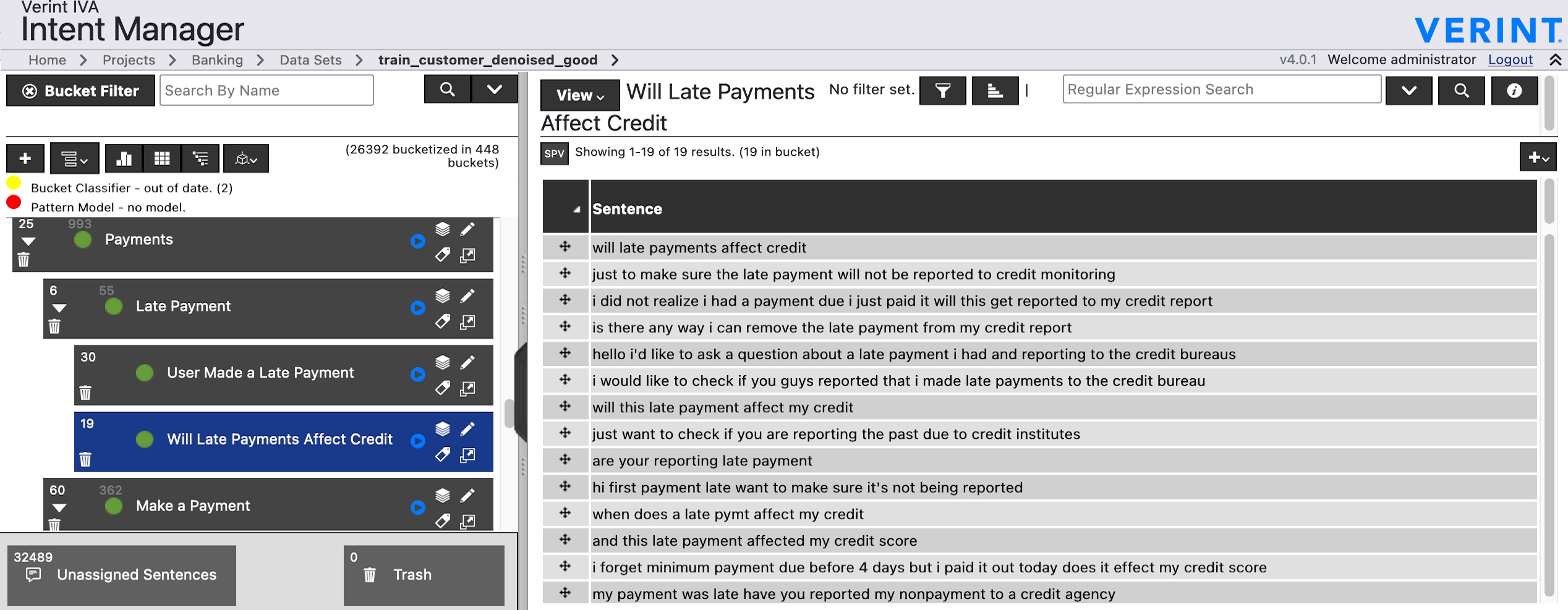}
  \vspace{-1mm}
  \caption{Work space of Verint Intent Manager. Analysts use this interface to organize texts into hierarchical intent labels.}
  \label{fig:teaser}
  \vspace{-3mm}
\end{figure*}

\section{Introduction}
Intelligent Virtual Assistants (IVAs) are becoming more popular in customer service and product support tasks~\cite{ram2018conversational}.  At Verint, the process to design and refine IVAs relies on data analysts who are familiar with specific terminology in a given language domain such as transportation or finance to mine customer service texts for latent user intentions, or \textit{intents}.  An intent is the interpretation of user input that allows the IVA to formulate the 'best' possible response and its detection is typically treated as a supervised classification problem in commercial IVAs~\cite{beaver2020automated}.  It is common for these analysts to receive a large batch of customer service logs from a company who would like to deploy an IVA to help automate some aspect of their customer service.  The analysts then need to mine this text data to surface the most common intents and determine which use cases an IVA would be able to easily automate.  They would then recommend a subset of high value intents for a company-specific IVA implementation to be deployed on a website, mobile application, or phone support line.

\textit{Verint Intent Manager (VIM)} is a powerful language classification tool that helps our data analysts to quickly review and organize such large volumes of unlabeled conversational text into various intents. Figure~\ref{fig:teaser} shows the main VIM work space where analysts search, filter, and group unlabeled text from customer service interactions (right table) and assign them to labels, then organize the labels into hierarchies (left sidebar)\footnote{A video walk through given by a data analyst of an older beta version of VIM, then named Prompt AI, is available here: ~\url{https://youtu.be/UjLu1L-ES0w}}.  These labeled texts are then exported to build intent classifier models for production IVAs.  As a manual investigative process is slow, using text clustering algorithms to provide recommendations for analysts to gain a better understanding of the type of language contained in input texts can partially automate the intent categorization process. We apply language models such as BERT, RoBERTa, and ELMo~\cite{peters2018deep, devlin2018bert, liu2019roberta} to encode texts to contextualized word representations for clustering in VIM.  However, there are three challenges that affect the performance of text clustering tasks on real-world customer service datasets.

First, the desired grouping of texts for describing user intentions may be different from those learnt by fully unsupervised approaches~\cite{wang2016semi}. Unless given further guidance, the contextualized text representations from pre-trained language models are more suited for finding latent topic clusters than surfacing user intentions~\cite{9377810}. Second, the size of unlabeled conversational logs obtained from contact centers varies from thousands to millions. The performance of the clustering method needs easily scale to such large volumes so that analysts can have the clustering results fast enough to iterate between labeling results and re-running the clustering on the remaining unlabeled pool. Third, the true number of clusters on a new dataset is often unknown and analysts may get sub-optimal results when using an inaccurate number of clusters.  Analysts then waste valuable time sorting through poor clustering results which defeats the purpose of automation.

To address these challenges for data exploration in the VIM platform,  we  developed an end-to-end text clustering pipeline that integrates four highly successful AI/ML techniques: transfer learning with deep language models (e.g., BERT), a high performance library for k-nearest neighbor (k-NN) graph construction and k-means clustering leveraging multiple GPUs, and a fast community detection algorithm to uncover high quality community structures. In this paper we discuss how we overcame these three challenges in our production application using this pipeline.

\begin{figure} [h]
\hspace{0.5cm}
\includegraphics[width=0.85\linewidth]{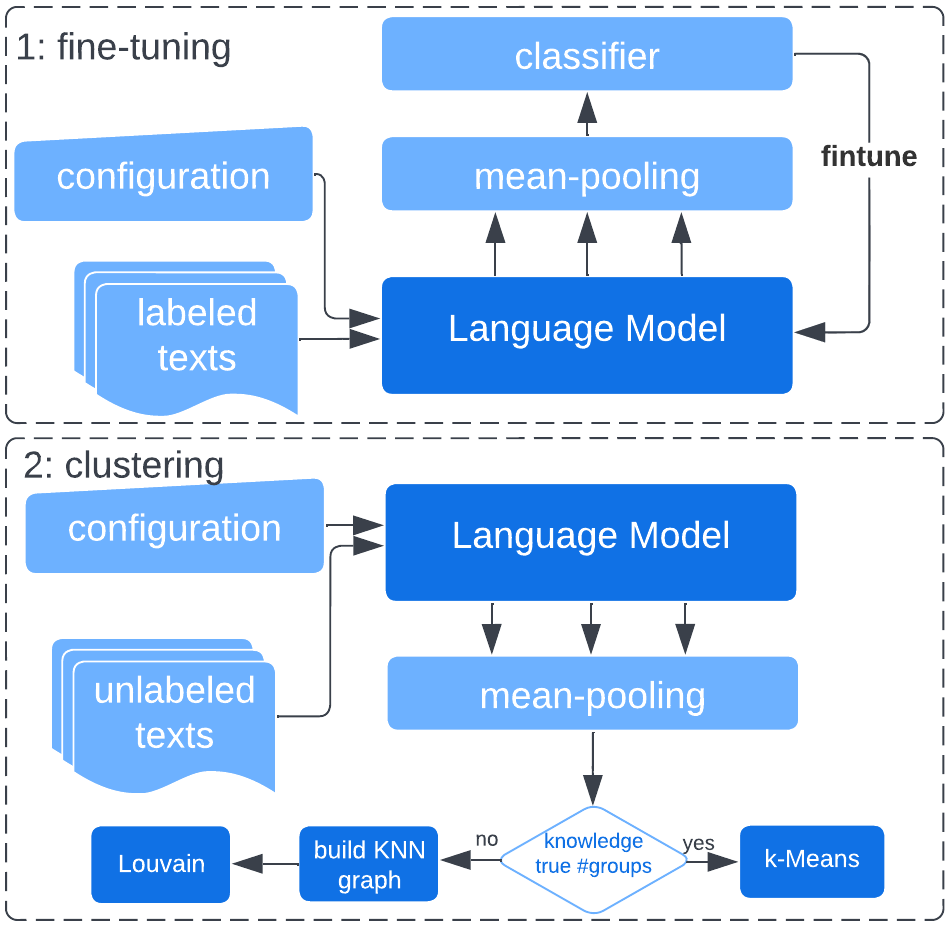}
\vspace{-1mm}
\caption{ Top: Fine-tuning a language model; 
Bottom: Embedding and flexible clustering pipeline. 
}
\label{fig:pipeline}
\vspace{-3mm}
\end{figure}

\section{The Pipeline}
Figure~\ref{fig:pipeline} shows our pipeline for flexible text clustering\footnote{Source code for the pipeline and additional experiments are made available here: \url{https://nextit-public.s3.us-west-2.amazonaws.com/FinetuneLM-Clustering.zip}}.
The system consists of the fine-tuning module and the flexible clustering module.  In the initial state when no data has been labelled yet, the untuned language model representations are directly used for downstream clustering as a fully unsupervised approach.  As the analyst works within the VIM interface (see Figure~\ref{fig:teaser}) they begin to organize and label some of the texts from the unsupervised clustering result or from manually created regex patterns.  Once the analyst has provided some small subset of labeled samples it is possible to fine-tune the language model to their particular task.  To fine-tune a language model, we add a linear classifier on top of the mean pooling layer. We optimise the cross entropy loss function to update the language model's parameters and the classifier at the same time.  After this fine-tuning stage, we discard the linear classifier and use the updated language model and mean pooling to produce text representations.

If the user does not specify the desired number of clusters, the representations from the language model are used to generate a k-NN graph using the \texttt{faiss} library~\cite{johnson2019billion}. The \texttt{faiss} library can efficiently provide similarity search and clustering with GPU support. 
Even with several millions of embedded sentence vectors, we can construct the k-NN graph by using exact searching with L2 distances in a few seconds on GPUs.

We use the k-means implementation of the \texttt{faiss} library or Louvain~\cite{blondel2008fast} depending on if the user specifies the desired number of clusters.  The Louvain algorithm can detect hierarchical clustering structures from large networks with millions of nodes and billions of links. Both algorithms can quickly handle the volumes of text commonly worked on by VIM users. The clustering results are presented to analysts (see screenshots in Figure~\ref{fig:squad-auto-cluster}) where they can click on a cluster to see member texts and bulk assign an intention label to a cluster, manually edit cluster membership, or launch a sub-clustering task to create intent hierarchies.  As the analyst iterates between clustering and labeling, the language model is being fine-tuned using their current labeled set creating a system that adapts to how each analyst is intending their dataset to be labeled.

\section{Analysis And Discussion}
In this section, we show the clustering results and try to answer the following questions regarding to our pipeline design: (1) How accurately can the pipeline separate different groups of texts? (2) Can Louvain detect high quality clusters without the data analyst knowing the true number in advance? (3) How fast can the pipeline complete end-to-end clustering tasks on different sized datasets?

\begin{table}
\centering
\small
\begin{tabular}{p{0.8cm} r p{1.37cm} p{1.67cm} p{1.67cm}}
\textbf{Dataset} &  \textbf{Size} & \textbf{\# Clusters} & \textbf{Purity}& \textbf{NMI} \\
\toprule
\textbf{RCV1} & 674k & 20.6 (4.0) & 0.913 (0.944) &  0.762 (0.808)  \\
 \midrule
\textbf{STKO} & 90k & 35.4 (10.0) & 0.926 (0.926) & 0.737 (0.823)   \\
 \midrule
\textbf{RCT} & 270k & 10.4 (5.0) & 0.787 (0.830) & 0.524 (0.608)  \\
\bottomrule
\end{tabular}
\vspace{-1mm}
\caption{
Compare clustering quality of Louvain and (k-Means). 
}
\label{T3}
\vspace{-3mm}
\end{table}

\subsection{Clustering Accuracy}


Clustering Purity and Normalized Mutual Information (NMI) are commonly used metrics for cluster quality evaluation~\cite{manning2008introduction}, both range from $0.0$ to $1.0$.  Higher purity means most texts belong to one true class in each detected cluster.  Higher NMI means good correlation between predicted cluster labels and true classes.  NMI penalizes splitting data into a large number of small clusters to correct this shortcoming of the purity score.  In Table~\ref{T3} we show an excerpt of three out of six public text datasets from an extensive evaluation of the pipeline we published in a previous work~\cite{9377810}.  These numbers were achieved by fine-tuning BERT with $2.5\%$ labeled samples of each dataset. In our additional experiments the pipeline achieved high clustering quality with both Louvain and k-means, and BERT and RoBERTa performed similarly but both outperformed ELMo.

\subsection{Deep Analysis of Louvain}
In Table~\ref{T3} we showed the number of clusters detected by Louvain is usually greater than the true number of clusters, shown in parentheses. However, the high purity scores indicates the majority of documents within the same predicted clusters are from the same true classes. Because documents can have hierarchical topics, Louvain appears to be detecting more fine-grained cluster structures inside each high level group from the k-NN graph.

\begin{table*}[t]
\centering
\small
\begin{tabular}{l|lllll}
\multicolumn{6}{c}{\textbf{Top-5 Frequent Bigrams for the RCV1-GCAT Cluster by Clustering Method}} \\
\toprule
\textbf{K-means} & & & \textbf{Louvain} & \\
\textbf{155,825}    & \textbf{735}& \textbf{21,589} & \textbf{736} & \textbf{98,863} & \textbf{2,047} \\
\midrule
\textbf{prime minister}   & official journal &  press digest & radio romania & prime minister & research institute \\
\textbf{press digest}  & journal contents &  stories vouch & vouch accuracy & hong kong & percent points \\
\textbf{stories vouch}   & contents oj &  verified stories & main headlines & united states & threegrade rating \\
\textbf{verified stories}   & note contents &  reuters verified  & radio headlines & foreign minister & rating system\\
\textbf{reuters verified}   & reverse order &  leading stories  & noon radio & news agency & whose values \\
\textbf{}  &\textbf{1,839} & \textbf{10,698} & \textbf{17,962} & \textbf{4,066} &  \\
\textbf{}  & emergency weather & world cup &  world cup & first innings &  \\
\textbf{ }  & weather conditions & first round &  first division & west indies &  \\
\textbf{}  & conditions wsc & grand prix &  w l & south africa &  \\
\textbf{}  & tropical storm & south africa &  1 0 & sri lanka &  \\
\textbf{}  & top winds & davis cup &  1 1 & new zealand &  \\
\bottomrule
\end{tabular}
\vspace{-1mm}
\caption{
 Clusters detected by Louvain and k-means from the government-and-social topic in RCV1. }
\label{T4}
\vspace{-3mm}
\end{table*}

In Table~\ref{T4}, the top-5 frequent bigrams from the 9 clusters detected by Louvain are compared with the bigrams from the k-means cluster corresponding to the topic category of GCAT (government and social) from the Reuters News (RCV1) dataset.  One example is $736$ news articles forms a separate group about Romania; $1,839$ articles forms another group about weather. These two clusters consist of only $0.4\%$ and $1.1\%$ of the articles in the entire GCAT category respectively. Another example is the separation of news related to two different \textit{world cup} events. By seeing the surfaced bigrams, analysts can understand the first group is more likely about tennis games and the second group is about soccer games\footnote{The \texttt{w 1} is a score like \texttt{Apr 9 Ajax H W 1-0}}. The two takeaways are: the k-NN graph accurately connects articles with their nearest neighbors; second, Louvain accurately detects communities with high resolution. This means VIM users get good clustering performance even if they do not know or specify the number of clusters for their task.

\vspace{-2mm}
\subsection{Computation Time}

We experiment with the running time of the pipeline across different Amazon EC2 instance types. Using a p3.16xlarge instance (eight V100 GPUs), the end-to-end clustering running time is 5.9x faster than on the p3.2xlarge instance (one V100 GPU).  When the pipeline is deployed on Amazon p3.16xlarge instances, the end-to-end computation can be finished under 5 minutes for the average task sizes, taking closer to 15 minutes for the largest dataset (673k texts).  This is acceptable for our use cases as analysts only run the pipeline periodically and then spend more time analyzing and labeling the result.  Detailed results are in the Appendix.




\begin{figure}[h!]
\hspace{0.5cm}
\includegraphics[scale=0.225]{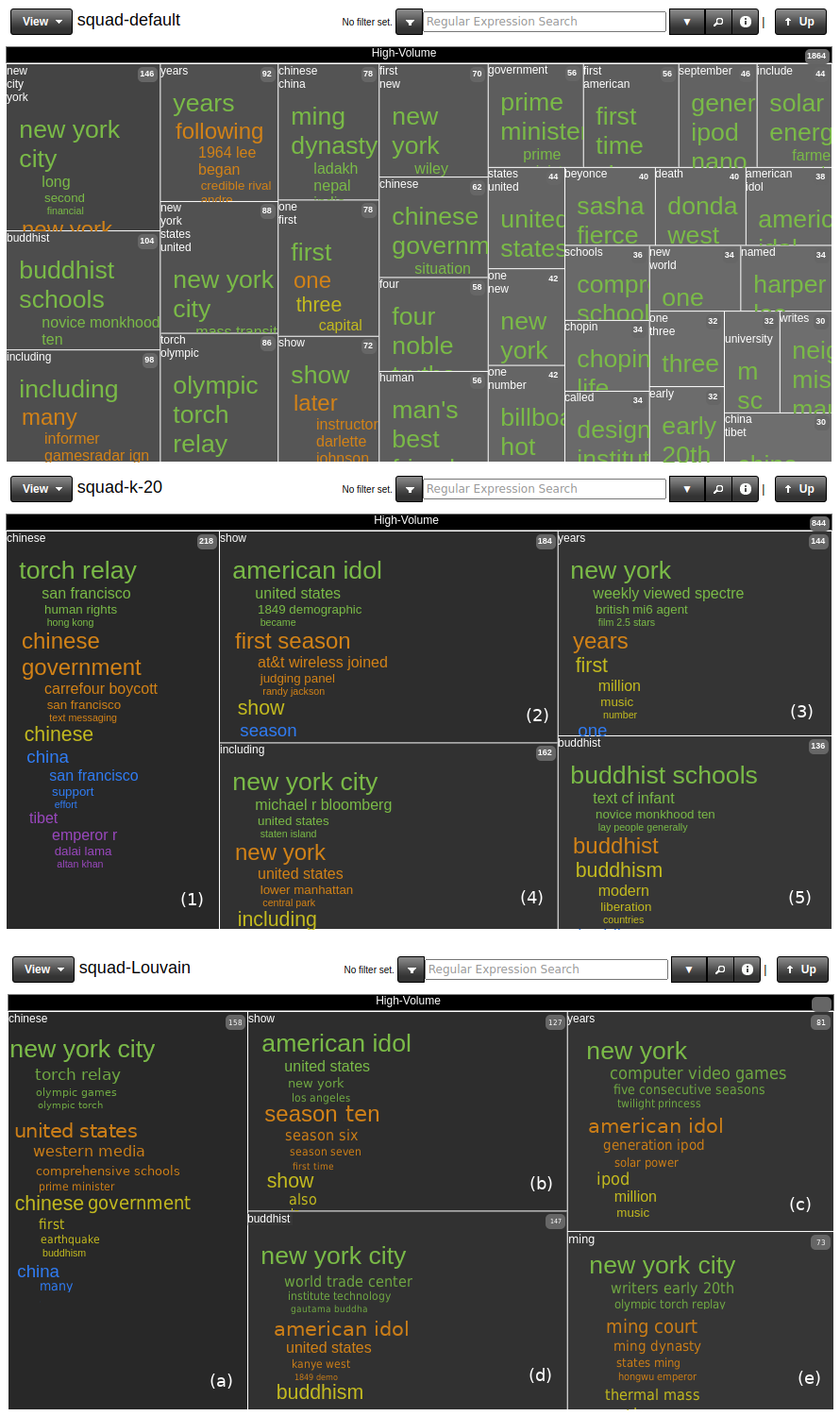}
\vspace{-2mm}
\caption{Frequent terms using (top) the default setting $k=250$.  (middle) ground truth $k=20$. (bottom) k-NN + Louvain to detect the number of groups $k=14$. }
\label{fig:squad-auto-cluster}
\vspace{-3mm}
\end{figure}
\vspace{-2mm}
\section{Case Study}
We use a case study to show our pipeline provides similar insight to analysts unfamiliar with a dataset as can be discovered by an analyst who is. We demonstrate the clustering pipeline has improved the usability of the VIM application even without a fine-tuning step.  When an analyst has no prior knowledge about the data the desired number of clusters is unknown.  Before adding our flexible clustering pipeline, in order to explore the data the analyst would either guess $k$ or use the default value.  The data for this study consists of 20 randomly selected articles from the SQUAD 2.0 dataset~\cite{rajpurkar2018know} where we extracted 1,400 paragraphs as the input texts to be labeled in VIM.

The top panel of Figure~\ref{fig:squad-auto-cluster} represents the scenario when an analyst has no prior knowledge about the data, as is the case with a new customer, so the number of intention groups is unknown. Using the default setting of k-means and $k=250$ (meant for larger datasets), VIM displays the top frequent terms selected from the $33$ largest clusters. Frequent terms appear in multiple groups without corresponding contextual terms, such as \textit{one, three, first time, new york}, making it difficult to gain insight from such clusters.

The middle panel shows the second scenario when the analyst is familiar with the data and he or she estimates the number of groups to be the true number $k=20$.  The VIM pipeline completes the clustering task using k-means and displays the frequent terms from the top $5$ largest clusters. Although there are common frequent terms in multiple clusters, other contextual terms and words that were important to each cluster surface so that analysts can at least identify three meaningful topic groups (Chinese government and the Olympic Games, American Idol show and Buddhism).

The bottom panel shows the results when VIM uses Louvain for community detection. The pipeline detects 14 groups from the 1,400 paragraphs. The detected number of groups are a good estimation of the true groups. Without prior knowledge of the dataset an analyst can now get the clustering result that is similar to scenario two. In this example, we can still identify the Chinese government and Olympic Games, American Idol and Buddhist topics. 
\vspace{-2mm}
\section{Conclusion}
\label{sec:conclusion}

In this paper, we have presented and open-sourced an important component in a commercial data exploration and annotation tool useful for labeling text in a user-adaptive manner.  Our text clustering pipeline is flexible and provides optimal results with a minimum of user configuration (specify number of clusters or not, provide a labeled sample for fine-tuning or not). The pipeline provides high quality clustering results and it is able to scale vertically on increasing text data volume and horizontally on increasing computation resources. We also showed the community detection techniques work well for text mining tasks, especially useful when working with unfamiliar data. On our current production configuration, analysts can get insightful clustering results in less than 15 minutes for the largest task evaluated. The clustering pipeline has greatly reduced the amount of data discovery effort required by the human analysts in our production application.


\bibliographystyle{icml2022}
\bibliography{sample-sigconf}

\begin{thebibliography}{11}
\providecommand{\natexlab}[1]{#1}
\providecommand{\url}[1]{\texttt{#1}}
\expandafter\ifx\csname urlstyle\endcsname\relax
  \providecommand{\doi}[1]{doi: #1}\else
  \providecommand{\doi}{doi: \begingroup \urlstyle{rm}\Url}\fi

\bibitem[Beaver \& Mueen(2020)Beaver and Mueen]{beaver2020automated}
Beaver, I. and Mueen, A.
\newblock Automated conversation review to surface virtual assistant
  misunderstandings: Reducing cost and increasing privacy.
\newblock \emph{Proceedings of the AAAI Conference on Artificial Intelligence},
  34:\penalty0 13140--13147, 04 2020.
\newblock \doi{10.1609/aaai.v34i08.7017}.

\bibitem[Blondel et~al.(2008)Blondel, Guillaume, Lambiotte, and
  Lefebvre]{blondel2008fast}
Blondel, V.~D., Guillaume, J.-L., Lambiotte, R., and Lefebvre, E.
\newblock Fast unfolding of communities in large networks.
\newblock \emph{Journal of statistical mechanics: theory and experiment},
  2008\penalty0 (10):\penalty0 P10008, 2008.

\bibitem[Chen et~al.(2020)Chen, Beaver, and Freeman]{9377810}
Chen, X., Beaver, I., and Freeman, C.
\newblock Fine-tuning language models for semi-supervised text mining.
\newblock In \emph{2020 IEEE International Conference on Big Data (Big Data)},
  pp.\  3608--3617, 2020.
\newblock \doi{10.1109/BigData50022.2020.9377810}.

\bibitem[Devlin et~al.(2018)Devlin, Chang, Lee, and Toutanova]{devlin2018bert}
Devlin, J., Chang, M.-W., Lee, K., and Toutanova, K.
\newblock Bert: Pre-training of deep bidirectional transformers for language
  understanding.
\newblock \emph{arXiv preprint arXiv:1810.04805}, 2018.

\bibitem[Johnson et~al.(2019)Johnson, Douze, and J{\'e}gou]{johnson2019billion}
Johnson, J., Douze, M., and J{\'e}gou, H.
\newblock Billion-scale similarity search with gpus.
\newblock \emph{IEEE Transactions on Big Data}, 2019.

\bibitem[Liu et~al.(2019)Liu, Ott, Goyal, Du, Joshi, Chen, Levy, Lewis,
  Zettlemoyer, and Stoyanov]{liu2019roberta}
Liu, Y., Ott, M., Goyal, N., Du, J., Joshi, M., Chen, D., Levy, O., Lewis, M.,
  Zettlemoyer, L., and Stoyanov, V.
\newblock Roberta: A robustly optimized bert pretraining approach.
\newblock \emph{arXiv preprint arXiv:1907.11692}, 2019.

\bibitem[Manning et~al.(2008)Manning, Raghavan, and
  Sch{\"u}tze]{manning2008introduction}
Manning, C.~D., Raghavan, P., and Sch{\"u}tze, H.
\newblock \emph{Introduction to information retrieval}.
\newblock Cambridge university press, 2008.

\bibitem[Peters et~al.(2018)Peters, Neumann, Iyyer, Gardner, Clark, Lee, and
  Zettlemoyer]{peters2018deep}
Peters, M.~E., Neumann, M., Iyyer, M., Gardner, M., Clark, C., Lee, K., and
  Zettlemoyer, L.
\newblock Deep contextualized word representations.
\newblock \emph{arXiv preprint arXiv:1802.05365}, 2018.

\bibitem[Rajpurkar et~al.(2018)Rajpurkar, Jia, and Liang]{rajpurkar2018know}
Rajpurkar, P., Jia, R., and Liang, P.
\newblock Know what you don't know: Unanswerable questions for squad.
\newblock \emph{arXiv preprint arXiv:1806.03822}, 2018.

\bibitem[Ram et~al.(2018)Ram, Prasad, Khatri, Venkatesh, Gabriel, Liu, Nunn,
  Hedayatnia, Cheng, Nagar, et~al.]{ram2018conversational}
Ram, A., Prasad, R., Khatri, C., Venkatesh, A., Gabriel, R., Liu, Q., Nunn, J.,
  Hedayatnia, B., Cheng, M., Nagar, A., et~al.
\newblock Conversational ai: The science behind the alexa prize.
\newblock \emph{arXiv preprint arXiv:1801.03604}, 2018.

\bibitem[Wang et~al.(2016)Wang, Mi, and Ittycheriah]{wang2016semi}
Wang, Z., Mi, H., and Ittycheriah, A.
\newblock Semi-supervised clustering for short text via deep representation
  learning.
\newblock \emph{CoNLL 2016}, pp.\ ~31, 2016.

\end{thebibliography}

\newpage
\appendix
\onecolumn
\section{End-to-End Clustering Time On AWS.}

\begin{figure}[h]
\centering
\includegraphics[scale=0.7]{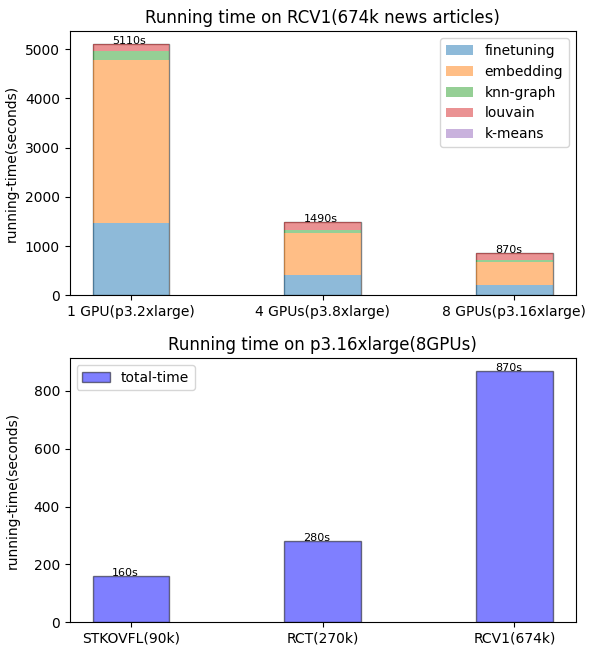}

\caption{\centering{(top) Running time of RCV1 data on different EC2 instances.  (bottom) End-to-end time of three datasets on the p3.16xlarge instance. }
}
\label{fig:runtime}
\end{figure}

The pipeline is scalable when we increasing the computation resources. The top plot in Figure \ref{fig:runtime} shows the pipeline gets $5.9$x speedup in end-to-end time when using 8 GPUs compared with using 1 GPU. The bottom plot shows the end-to-end time grows sub-linearly when the size of task dataset increases (Note the x-axis is not scaled in proportion to dataset sizes). The largest task (i.e., 674k news articles RCV1) can finish in $14.5$ minutes on the p3.16xlarge instance (using 8 V100 GPUs).  Of the three tasks, STKO has over 90,000 question texts, which would be near the average dataset size for the users of VIM that we have observed in production so far. The RCT task resembles a larger dataset size analysts do encounter quite often. The RCV1 task contains 0.67 million news articles, which is analogous to a large production work load we have observed in VIM.


\end{document}